\documentclass{article}
\usepackage[utf8]{inputenc}
\usepackage{authblk}
\usepackage{setspace}
\usepackage[margin=1.25in]{geometry}
\usepackage{graphicx}
\graphicspath{ {./figures/} }
\usepackage{subcaption}
\usepackage{amsmath}
\usepackage{lineno}

\usepackage{framed,multirow}
\usepackage{amssymb, amsmath}
\usepackage{latexsym}
\usepackage{url}
\usepackage{xcolor}
\definecolor{newcolor}{rgb}{.8,.349,.1}

\usepackage{graphicx}
\usepackage{caption}
\usepackage{subcaption}
\usepackage{threeparttable}
\usepackage{algorithmic, algorithm2e}
\usepackage{paralist}
\usepackage{dblfloatfix}
\usepackage{array}
\newcolumntype{C}[1]{>{\centering\arraybackslash}p{#1}}
\usepackage{tablefootnote}
\usepackage{lineno}

\usepackage[style=nejm, 
citestyle=numeric-comp,
sorting=none]{biblatex}
\title{Analysis of Learned Features and Framework for Potato Disease Detection}
\author[1*$\dag$]{Shikha Gupta}
\author[2$\dag$]{Soma Chakraborty}
\author[1]{Renu Rameshan}

\affil[1]{Vehant Technologies Private Limited, Noida, Uttar Pradesh 201301 India.}
\affil[2]{Indian Institute of Technology, Mandi, HP-175005, India.}

\affil[*]{shikhag@vehant.com}
\affil[$\dag$]{These authors contributed equally to this work.}
\affil[1]{Both the authors were affiliated to IIT Mandi during the duration and completion of this work.}

\date{}

\begin{document}
\maketitle

\begin{abstract}
For applications like plant disease detection, usually, a model is trained on publicly available data and tested on field data. This means that the test data distribution is not the same as the training data distribution, which affects the classifier performance adversely. We handle this dataset shift by ensuring that the features are learned from disease spots in the leaf or healthy regions, as applicable. This is achieved using a faster Region-based convolutional neural network (RCNN) as one of the solutions and an attention-based network as the other. The average classification accuracies of these classifiers are approximately 95\%, while evaluated on the test set corresponding to their training dataset. These classifiers also performed equivalently with an average score 84\% on a dataset not seen during the training phase.

\end{abstract}
\textbf{Keywords:} Potato disease detection, Object detection-based approach, Attention-based network, Machine learning, Deep learning 
\section{Introduction}
Automatic disease detection software implemented in unmanned aerial vehicles (UAV) or hand-held mobile devices can be a crucial tool for the initial diagnosis of plant diseases in fields, especially in vast plantations where manual monitoring is not feasible or in remote locations with limited physical access. Plant diseases often manifest as visible spots on the leaves, which can be imaged easily using mobile cameras. Training machine learning models using these images offer a cost-effective approach to developing automated plant disease detection systems.

For these models, robustness and applicability to real-world plant images with guarantees happen only when the features are discriminative and learned from the region of an image containing disease spots. However, in our previous work \cite{myPaper}, it was observed that the environment of sample collection and image quality issues like uneven illumination, defocus, low inter-class variation, and low intra-class similarity of images, etc., significantly impact the extraction of desired features from the images.

In \cite{myPaper}, several convolutional neural network (CNN) based classification models were built using images obtained from unconstrained environments of potato plantation sites  (\textit{in-field} dataset), a laboratory setup (\textit{lab-prepared} dataset), and segmented images from in-field datasets. It was observed that the average testing performance of these classifiers was approximately 90\%, but the average cross-testing accuracy of all these models was below 50\%. \emph{Cross-testing refers to evaluating classifiers using test samples from datasets other than their respective training datasets}. Analyzing this inconsistent classification performance, it was inferred that the discriminative features learned by the models from their respective datasets were not precisely relevant to identifying disease spots and healthy parts of the leaves. These models were learning features from background shapes etc., which are not relevant. 

In this work, we qualitatively analyze the learned features using an advanced CNN visualization technique. Based on this analysis, we propose and provide experimental proof of best-suited classification models for plant disease detection using leaf images. The following is a summary of our contributions:

\begin{enumerate}
	
	\item Quantitative evaluations of class-discriminative salient regions to demonstrate that traditional CNNs tend to learn the broader contextual information rather than focusing specifically on disease spots and healthy regions of a leaf image.
	\item A region-based image classifier is proposed by adapting a convolutional neural network-based object detection framework using \textit{in-field} dataset. The testing accuracy of this classifier is 96.95\% with 91.85\% region overlap index, and cross-testing accuracy (using the \textit{lab-prepared} dataset) is 84.76\%.  
	
	\item An attention-based classification framework is proposed, giving a high response value for the image's disease spot(s) (attention location). The testing accuracy of this classifier is 94.17\%, and cross-testing accuracy (using the \textit{lab-prepared} dataset) is 81.65\%.

\end{enumerate}
The paper is organized as follows: Section \ref{sec:relWorks} contains the related works, Section \ref{sec:propApp} describes the proposed approach and results, Section \ref{sec:obsAnl} reports the observations and analysis, and Section \ref{sec:conc} contains the conclusions and possible future directions.
\section{{{Related Work}}} \label{sec:relWorks}
\vspace{-0.3cm}
\noindent We are reviewing only those works from the literature that deal with image-based automatic plant disease detection systems. Several methods are reported using various state-of-the-art machine learning models trained on images of different kinds of crop and vegetable plant leaf images \cite{Survey20}, \cite{Survey19}, and \cite{Survey18}. All these works mainly compared the performance of conventional machine learning models like SVM or some well-known CNN models like AlexNet \cite{alexNet}, VGG16 \cite{VGG16}, InceptionV3 \cite{inception}, FRCNN \cite{FRCNN}, SSD \cite{ssd}, Mask RCNN \cite{he2017mask} etc. Most of these methods used a \textit{lab-prepared} dataset called \textit{Plant Village} \cite{PlantVill16} and reported good classification performance (average accuracy is 96\%). Few works also reported accuracies ranging from 89\%  to 98\% using a few hundred to a few thousands of \textit{in-field} samples. Joe. et al. \cite{johnson2021enhanced} trained Mask RCNN-based instance segmentation model~\cite{he2017mask} using RGB and different color space images to detect the diseased patch. The reported mean average precision (mAP) and mean average recall (mAR) on RGB images are 80.9\% and 55.5\%, respectively. Additionally, they reported higher values of 98.1\% for mAP and 81.9\% for mAR, but those results are not taken into consideration due to the involvement of manual intervention in their calculation.

However, none of these studies provided confirmation regarding the classifier's adaptability and effectiveness when dealing with images of closely related plant species taken under various real-world conditions.
Also, the analysis and reasoning of varying performances by the proposed models were not reported. Methods using \textit{in-field} data also lack analysis and justification for achieving high accuracy with fewer real-world images. Few methods used well-known CNN visualization methods and reported the highlighted disease spots in \textit{lab-prepared} images \cite{Survey20}. But a quantitative measure of the impact of highlighted image regions on the model's performance has not been reported. Such works are not reviewed in this paper.

As stated earlier, the key factor in creating an efficient plant leaf image-based disease detection system is extracting discriminative and relevant features of disease spots and the leaves in the images. So, for such systems, quantitative and qualitative analysis of learned features must be emphasized. Moreover, the environment of collected samples and faults in captured images impact the classifier's performance significantly. For proper validation of the plant disease classifiers, experiments should be conducted using both \textit{lab-prepared} and \textit{in-field} images on models of the same architecture. 

\begin{table*}[h!]
	\fontsize{9}{11}\selectfont
	\caption{Description of the datasets used to train the classifiers in \cite{myPaper} and in the present work.}
	\label{table:clsfr_ds}
	\centering
	\begin{threeparttable}[t]
		\begin{tabular}{|c|p{0.7\linewidth}|}\hline
			
			{Dataset} & {Description} \\\hline
			
			\textbf{org-CPRI} (\textit{in-field}) & A dataset of images, captured in potato plantation fields in Northern India and is provided by the Central Potato Research Institute, Shimla, India (Figure \ref{fig:sampleImgs}(\subref{fig:org_cpri})). A detailed description of this dataset is in  \cite{myPaper}. \\\hline
			
			\textbf{seg-CPRI} & Segmented CPRI dataset generated by the trained UNet model \cite{ronneberger2015u} followed by post-processing steps using morphological operations (the second column in Figure \ref{fig:sampleImgs}(\subref{fig:seg_cpri})). \\\hline
			
			\textbf{PV} (\textit{lab-prepared}) & A public dataset of images captured in the constrained environment of a laboratory (Figure \ref{fig:sampleImgs}(\subref{fig:pv})). This dataset was created for the work \textit{Plant Village} \cite{PlantVill16}. \\\hline	\end{tabular}
	\end{threeparttable} 
\end{table*}

\section{Proposed Approach} \label{sec:propApp}

\begin{figure}
	\centering
	\begin{subfigure}[t]
		{\dimexpr0.13\textwidth+20pt\relax} 
		\makebox[20pt]{\raisebox{28pt}{\rotatebox[origin=c]{90}{\footnotesize{EB}}}}%
		\includegraphics[width=\dimexpr\linewidth-20pt\relax]
		{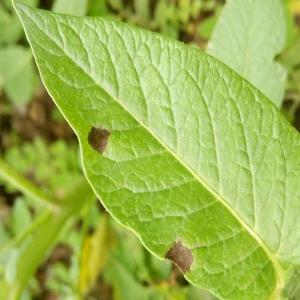}\vspace{0.1cm}
		\makebox[20pt]{\raisebox{28pt}{\rotatebox[origin=c]{90}{\footnotesize{LB}}}}%
		\includegraphics[width=\dimexpr\linewidth-20pt\relax]
		{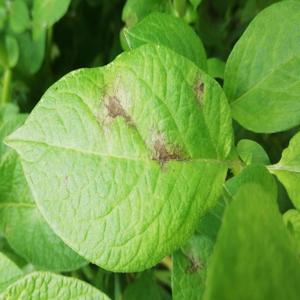}\vspace{0.1cm}
		\makebox[20pt]{\raisebox{28pt}{\rotatebox[origin=c]{90}{\footnotesize{HL}}}}%
		\includegraphics[width=\dimexpr\linewidth-20pt\relax]
		{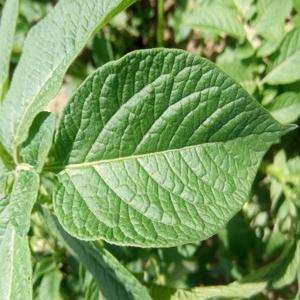}\vspace{0.1cm}

		\caption{org-CPRI}
		\label{fig:org_cpri}
	\end{subfigure}\hspace{0.1cm}
	\begin{subfigure}[t]{0.13\textwidth}
		\includegraphics[width=\textwidth]  
		{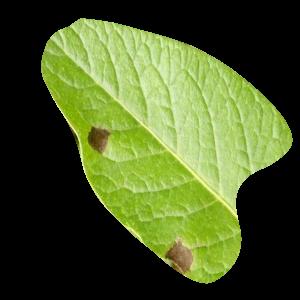}\vspace{0.1cm}
		\includegraphics[width=\textwidth]
		{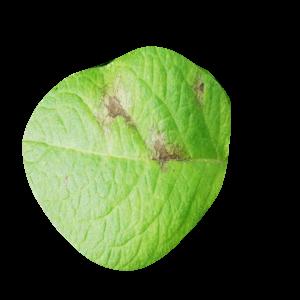}\vspace{0.05cm}
		\includegraphics[width=\textwidth]
		{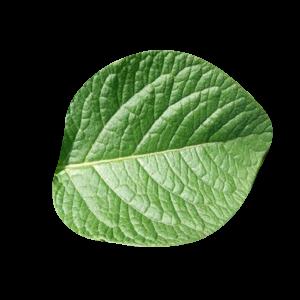}\vspace{0.05cm}
		\caption{seg-CPRI}
		\label{fig:seg_cpri}
	\end{subfigure}\hspace{0.1cm}
	\begin{subfigure}[t]{0.13\textwidth}
		\includegraphics[width=\textwidth]  
		{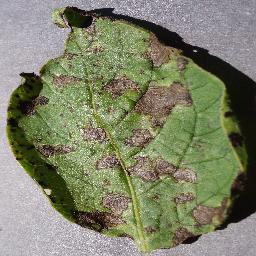}\vspace{0.1cm} 
		\includegraphics[width=\textwidth]  
		{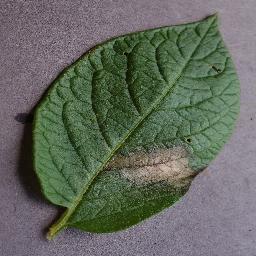}\vspace{0.05cm}
		\includegraphics[width=\textwidth]
		{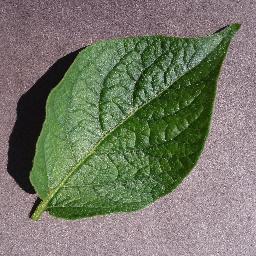}\vspace{0.05cm}	
		\caption{PV}
		\label{fig:pv}
	\end{subfigure}	
	\caption{Sample images from three used datasets: (a) original CPRI images (org-CPRI), (b) segmented CPI images (seg-CPRI), (c) Plant Village (PV) images. }
	\label{fig:sampleImgs}
\end{figure}

We start this work by analyzing the learned features of CNN-based models trained in \cite{myPaper} to classify potato leaf images into three classes - \textit{early blight} (EB), \textit{late blight} (LB) and \textit{healthy leaves} (HL) (Figure \ref{fig:sampleImgs}). Table \ref{table:clsfr_ds} contains a concise description of the datasets used. A state-of-the-art CNN visualization method is applied to the trained models for quantitative and qualitative analysis of the learned features, as described in sections \ref{ssec:vis} and \ref{ssec:qntv_measures}.

Quantitative measures of salient regions revealed that the features are learnt from the overall organization of image content and are not directly related to disease spots and healthy parts of leaves, i.e., there models failed to locate the region of interest in images to learn class discriminative features. An appropriate model for an effective classifier must learn to localize the region of interest in an image and should learn features specific to these regions.

We implemented such guided learning of features related to disease spots and healthy leaves in two ways: 
\begin{inparaenum}
	\item using an object detection-based framework - faster region-based convolutional neural network (Faster-RCNN) \cite{FRCNN}, and
	\item using an attention-based classification framework - attention branch network (ABN) \cite{fukui2019attention}.
\end{inparaenum} Both the methods are as described in section \ref{ssec:imCls}. 

\subsection{Salient Regions Visualization} \label{ssec:vis}
  	 	
\begin{figure}%
	\centering
	\begin{subfigure}[t]{0.30\textwidth}
		\includegraphics[width=\textwidth]{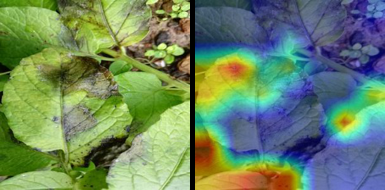}%
		\caption{Original CPRI}
		\label{fig:eb_ftb5}
	\end{subfigure}	
	\qquad
	\begin{subfigure}[t]{0.30\textwidth}
		\includegraphics[width=\textwidth]{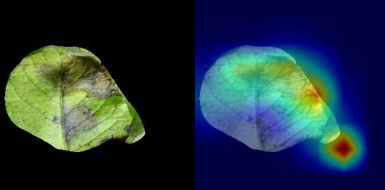}%
		\caption{Segmented CPRI}
		\label{fig:lb_ftb5}
	\end{subfigure}
	\qquad
	\begin{subfigure}[t]{0.30\textwidth}
		\includegraphics[width=\textwidth]{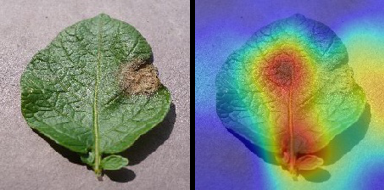}%
		\caption{PV}
		\label{fig:hl_ftb5}
	\end{subfigure}
	\caption{Maximally activated image regions (\textit{saliency maps}) generated by the classifiers trained on the three datasets using the class activation mapping method.}
	\label{fig:ClassSaliency_FTB5}
\end{figure}

The class decisive regions were visualized (Figure \ref{fig:ClassSaliency_FTB5}) using the work in \cite{Selvaraju17}. These visualizations (saliency maps) provide the coarse localization of the dominant parts of an image that are responsible for a class decision by a trained model. \emph{The red to orange regions in the figure are the most important regions for a class decision.} 

\subsection{Quantitative Measures of Salient Regions Generated by Trained Models}\label{ssec:qntv_measures}
  	
The correctness of saliency maps generated by the trained models is quantitatively measured with respect to the ground truth of CPRI and PV samples by calculating overlap precision and overlap recall:

  	\begin{equation}
overlapPrecision=\frac{TP}{TP+FP}
\end{equation}
 	\begin{equation}
  	 overlapRecall=\frac{TP}{TP+FN}  
   \end{equation}
  	where, $TP$ stands for \textit{true positive} pixels that are $1$ in both the ground-truth and saliency map. $FP$ refers to \textit{false positive} pixels that are $0$ in the ground-truth but $1$ in saliency map, and $FN$ refers to \textit{false negative} pixels that are $1$ in the ground-truth but $0$ in saliency map. Hence, \textit{overlapPrecision} measures the amount of true information in the saliency maps, and \textit{overlapRecall} measures the amount of ground-truth information that is retrieved by the saliency maps. Table \ref{table:metricsSalReg} presents the dataset-wise average metrics per class.
  	
\begin{figure}[t]
	\centering
	\begin{subfigure}[t]{\dimexpr0.13\textwidth+20pt\relax}
		\makebox[20pt]{\raisebox{20pt}{\rotatebox[origin=c]{90}{\footnotesize{PV}}}}%
		\includegraphics[width=\dimexpr\linewidth-20pt\relax]
		{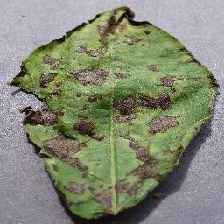}\vspace{0.05cm}
		\makebox[20pt]{\raisebox{20pt}{\rotatebox[origin=c]{90}{\footnotesize{orgCPRI}}}}%
		\includegraphics[width=\dimexpr\linewidth-20pt\relax]
		{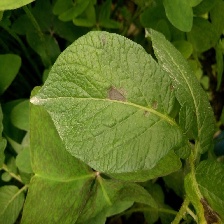}\vspace{0.05cm}
		\makebox[20pt]{\raisebox{20pt}{\rotatebox[origin=c]{90}{\footnotesize{segCPRI}}}}%
		\includegraphics[width=\dimexpr\linewidth-20pt\relax]
		{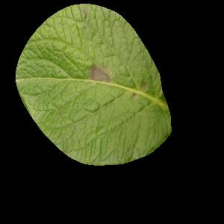}		
		\caption{Original}
		\label{fig:eb_org}
	\end{subfigure}
	\begin{subfigure}[t]{0.13\textwidth}
		\includegraphics[width=\textwidth] 
		{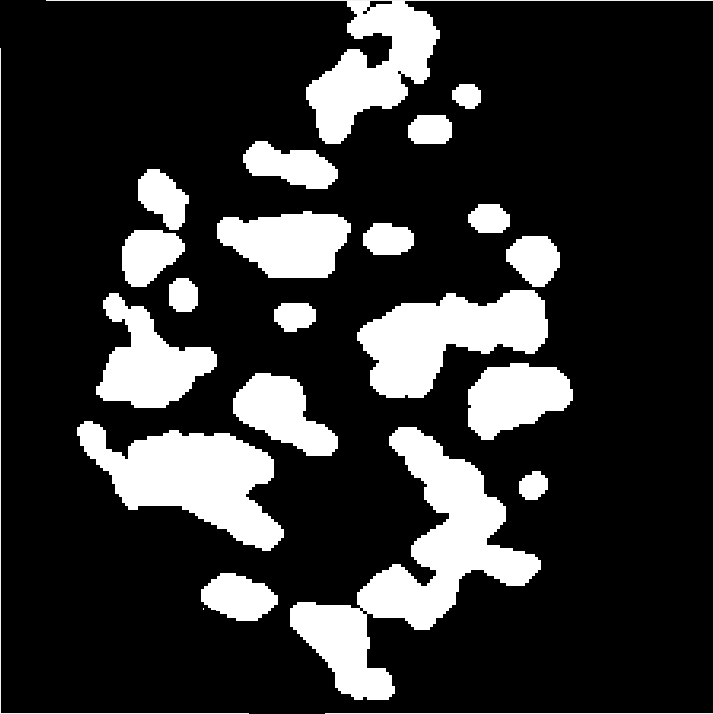}\vspace{0.05cm}
		\includegraphics[width=\textwidth]  
		{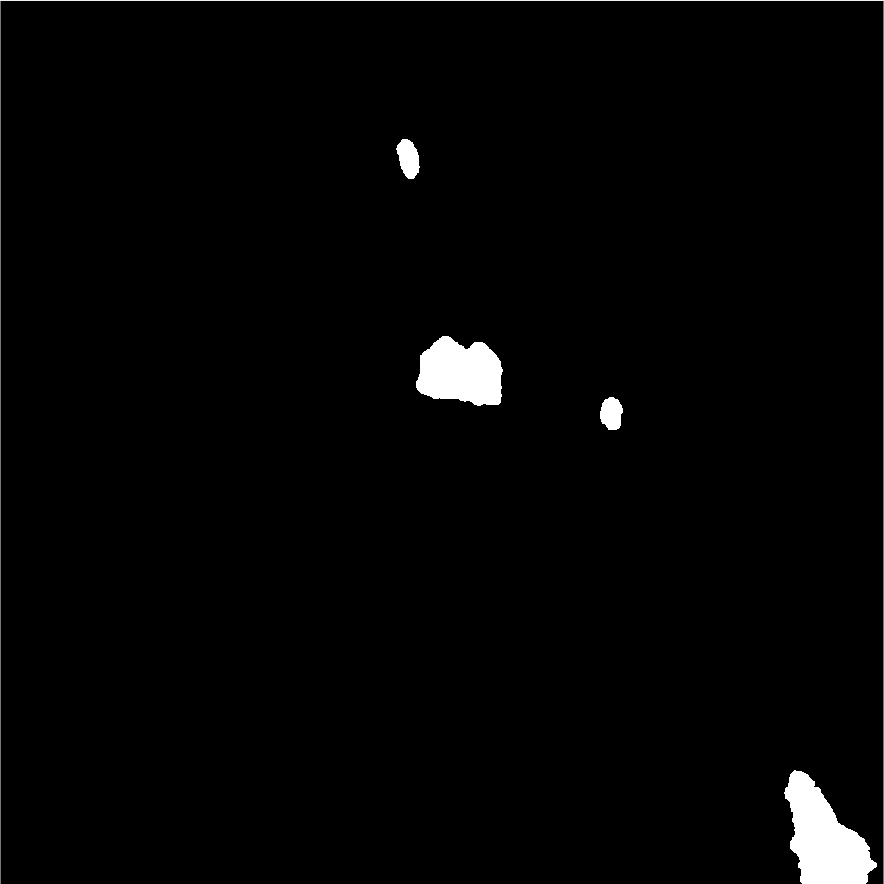}\vspace{0.05cm}
		\includegraphics[width=\textwidth] 
		{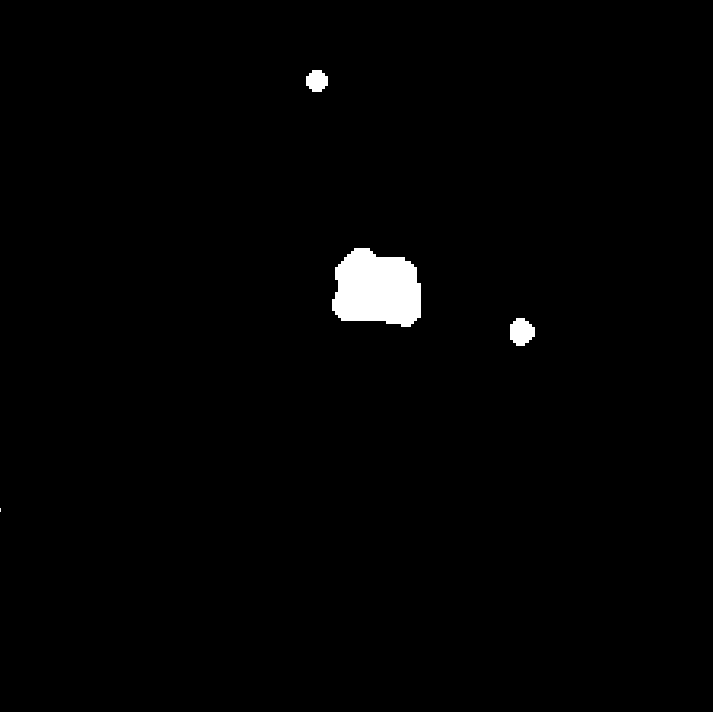}		
		\caption{GT}
		\label{fig:eb_gt}
	\end{subfigure}
	\begin{subfigure}[t]{0.13\textwidth}
		\includegraphics[width=\textwidth]   
		{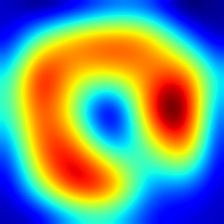}\vspace{0.05cm}
		\includegraphics[width=\textwidth]  
		{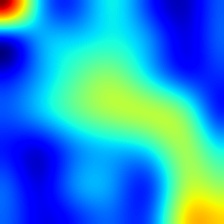}\vspace{0.05cm}
		\includegraphics[width=\textwidth]
		{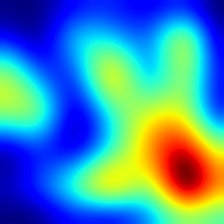}		
		\caption{SM}
		\label{fig:eb_sm}
	\end{subfigure}
	\begin{subfigure}[t]{0.13\textwidth}
		\includegraphics[width=\textwidth]  
		{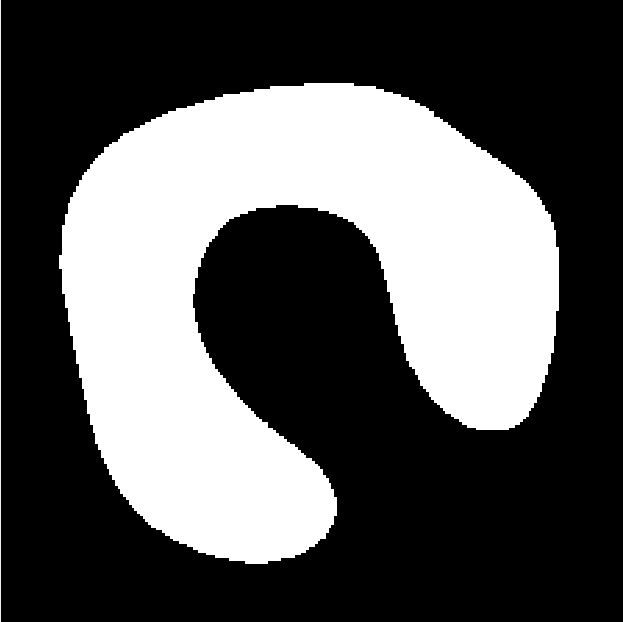}\vspace{0.05cm}
		\includegraphics[width=\textwidth]  
		{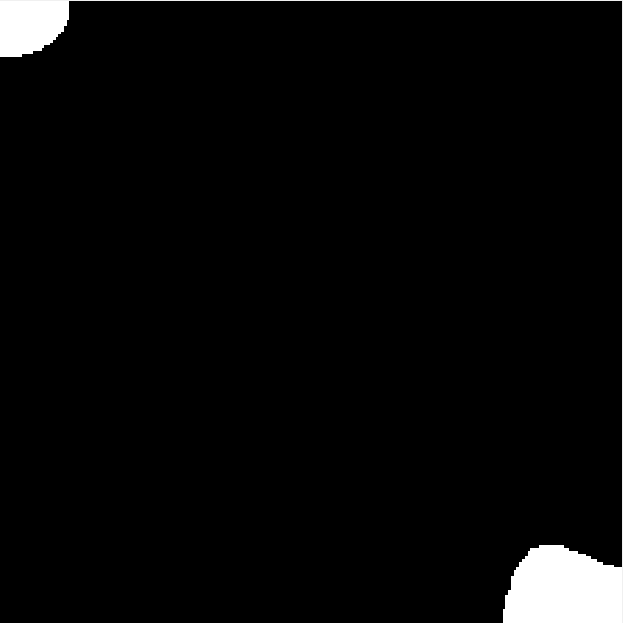}\vspace{0.05cm}
		\includegraphics[width=\textwidth]
		{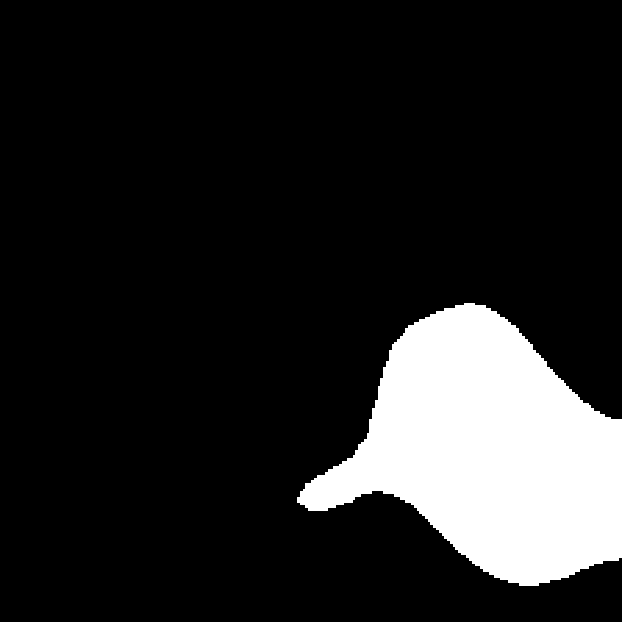}		
		\caption{BSM}
		\label{fig:eb_bsm}
	\end{subfigure}

	\caption{Examples of the color-coded ground-truth images (GT), saliency maps (SM), and binarised saliency maps (BSM) of \textit{early blight} (EB)  class.}
	\label{fig:quantify_salReg_eb}
\end{figure}

\begin{figure}[t]
		\centering
		\begin{subfigure}[t]{\dimexpr0.13\textwidth+20pt\relax}
			\makebox[20pt]{\raisebox{20pt}{\rotatebox[origin=c]{90}{\footnotesize{PV}}}}%
			\includegraphics[width=\dimexpr\linewidth-20pt\relax]
			{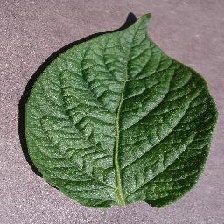}\vspace{0.07cm}
			\makebox[20pt]{\raisebox{20pt}{\rotatebox[origin=c]{90}{\footnotesize{org-CPRI}}}}%
			\includegraphics[width=\dimexpr\linewidth-20pt\relax]
			{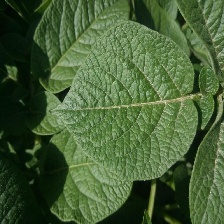}\vspace{0.07cm}
			\makebox[20pt]{\raisebox{25pt}{\rotatebox[origin=c]{90}{\footnotesize{seg-CPRI}}}}%
			\includegraphics[width=\dimexpr\linewidth-20pt\relax]
			{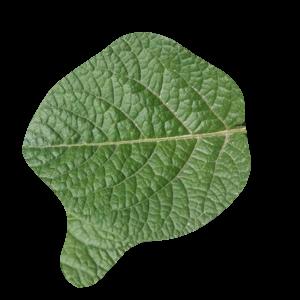}\vspace{0.07cm}			
			\caption{Original}
			\label{fig:hl_org}
		\end{subfigure}
		\begin{subfigure}[t]{0.13\textwidth}
			\includegraphics[width=\textwidth]
			{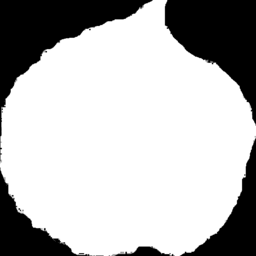}\vspace{0.07cm}
			\includegraphics[width=\textwidth]  
			{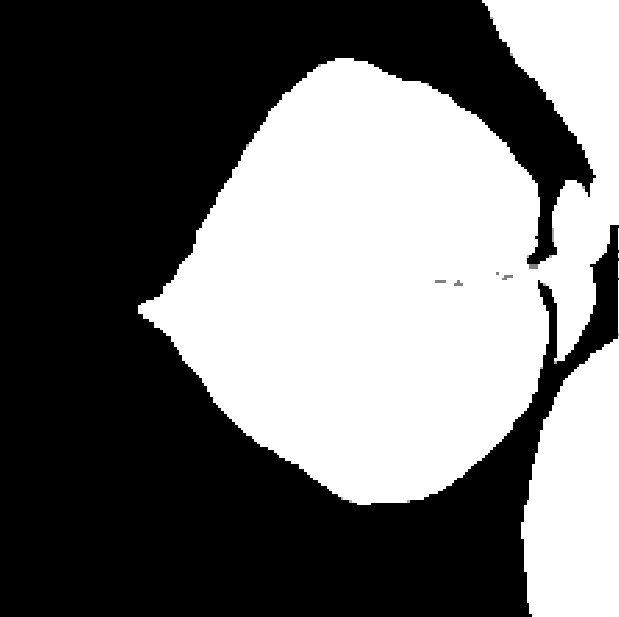}\vspace{0.07cm}
			\includegraphics[width=\textwidth]  
			{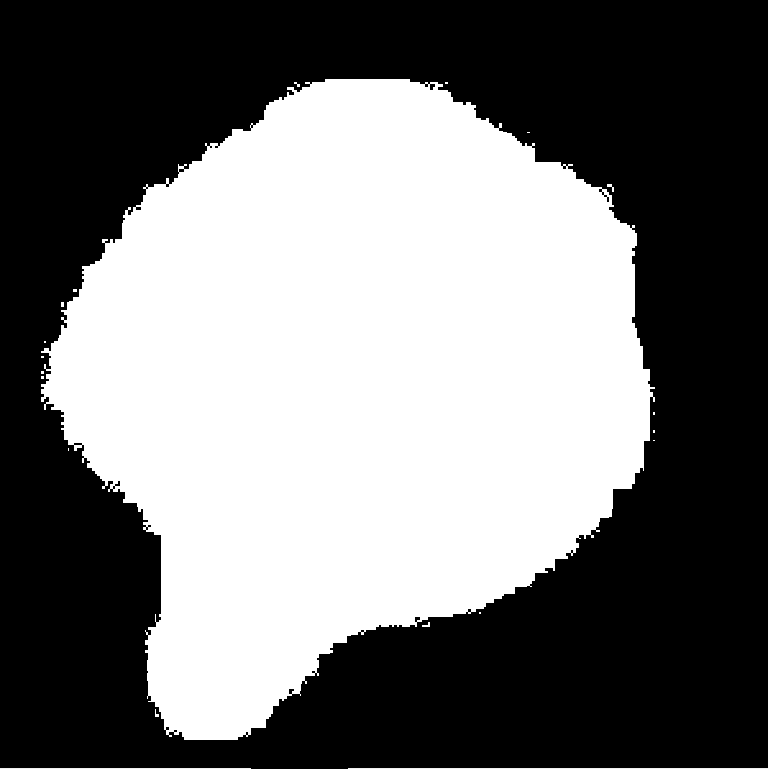}			
			\caption{GT}
			\label{fig:hl_gt}
		\end{subfigure}
		\begin{subfigure}[t]{0.13\textwidth}
			\includegraphics[width=\textwidth]  
			{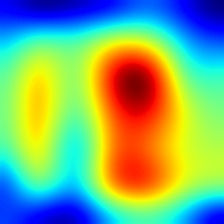}\vspace{0.07cm}
			\includegraphics[width=\textwidth]  
			{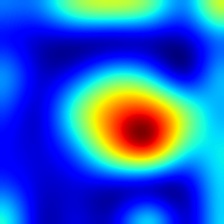}\vspace{0.07cm}
			\includegraphics[width=\textwidth]  
			{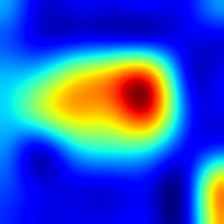}		
			\caption{SM}
			\label{fig:hl_sm}
		\end{subfigure}
		\begin{subfigure}[t]{0.13\textwidth}
			\includegraphics[width=\textwidth]
			{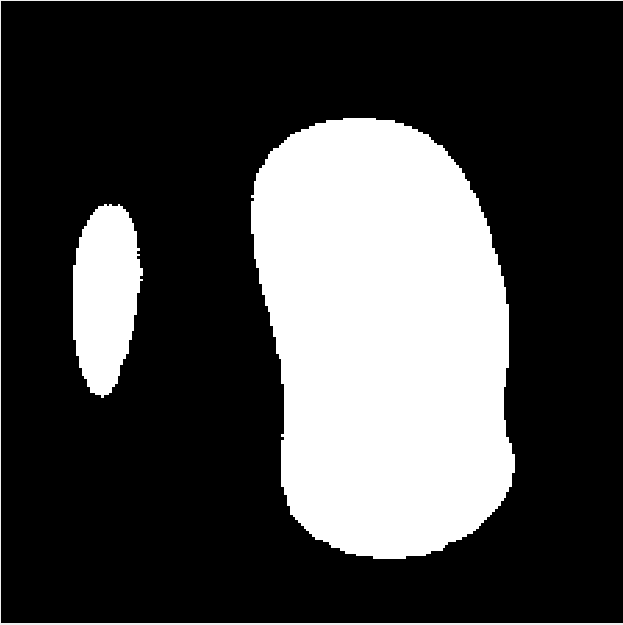}\vspace{0.07cm}
			\includegraphics[width=\textwidth]  
			{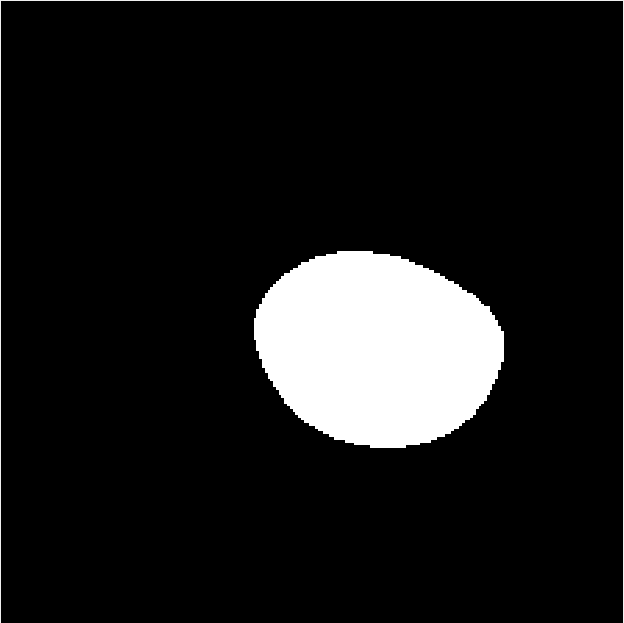}\vspace{0.07cm}
			\includegraphics[width=\textwidth]  
			{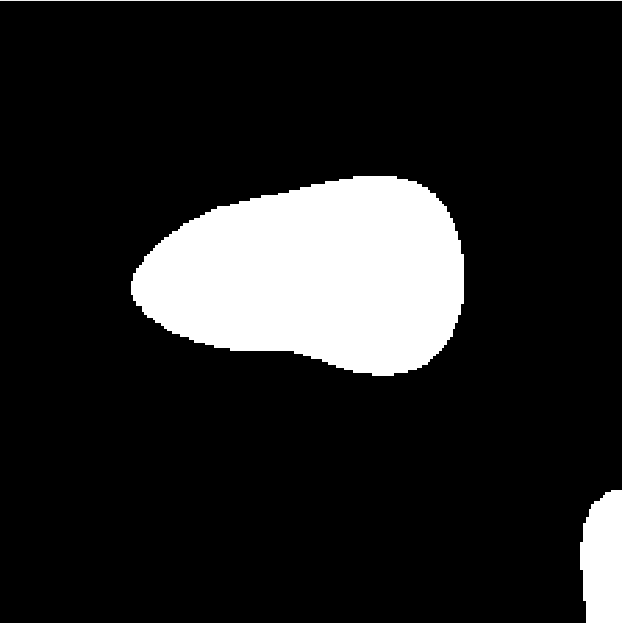}		
			\caption{BSM}
			\label{fig:hl_bsm}
		\end{subfigure}		
		
		\caption{Example of the color-coded ground-truth images (GT), saliency maps (SM), and binarized saliency maps (BSM) of \textit{healthy leaves} (HL) class.}
		\label{fig:quantify_salReg_hl}
	\end{figure}

  	The foregrounds of the disease class images are disease spots surrounded by a narrow strip of the healthy part of the leaf, and foregrounds of \textit{healthy leaves} class contain one prominent leaf. The ground-truth images are binarized by thresholding the green index of the hue channel of HSV colorspace (Figure \ref{fig:eb_gt} and Figure \ref{fig:hl_gt}). The threshold for healthy leaf is between 0.15 to 0.60, and it is less than 0.15 for disease spots. The saliency maps are binarised by setting pixels in red to orange colored regions to 1 (Figure \ref{fig:eb_bsm} and Figure \ref{fig:hl_bsm}).

\begin{table*}[t]
	\fontsize{9}{11}\selectfont
	\caption{Evaluation metrics of saliency maps generated by the trained models in \cite{myPaper}}
	\label{table:metricsSalReg}
	\centering
	\begin{tabular}{|c|C{0.07\textwidth}|C{0.07\textwidth}|C{0.07\textwidth}|C{0.07\textwidth}|C{0.07\textwidth}|C{0.07\textwidth}|C{0.07\textwidth}|C{0.07\textwidth}|}\hline
		\multirow{8}{*}{}  \textbf{Dataset} & \multicolumn{3}{|c|}{\textbf{Overlap Precision} (\%)} & \textbf{\textit{Avg. Precision}} &  \multicolumn{3}{|c|}{\textbf{Overlap Recall (\%)}} & \textbf{\textit{Avg. Recall}} \\\hline
		& {\textbf{Early Blight}} & {\textbf{Late Blight}} & {\textbf{Healthy Leaves}} & & {\textbf{Early Blight}} & {\textbf{Late Blight}} & {\textbf{Healthy Leaves}} & \\ \hline
		{PV} & 66.17 & 49.92 & 86.80 & \textit{67.63} & 30.50 & 53.03 & 44.47 & \textit{42.67} \\\hline
		{org-CPRI} & 12.95 & 10.65 & 68.57 & \textit{30.72} & 24.85 & 25.33 & 24.87 & \textit{25.12} \\ \hline
		{seg-CPRI} & 33.05 & 29.87 & 90.71 & \textit{74.64} & 17.47 & 15.46 & 24.24 & \textit{57.17} \\ \hline
		\hline
		\textbf{Avg.} & & & & \textbf{49.92} & & & & \textbf{28.91} \\ \hline

	\end{tabular}
\end{table*}

\subsection{ROI Based Image Classification:} \label{ssec:imCls}

This classifier comprises two parts - \begin{inparaenum}[(I)]
	\item ROI detector and
	\item detected ROI-based image classifier. 
\end{inparaenum} \emph{ROI is a bounding box around a diseased spot or a whole, healthy leaf in an image}.

\begin{enumerate}[I]
	\item \textbf{ROI Detector: }
A trainable part to locate disease spots or healthy leaf in an image and to predict their class labels simultaneously, implemented using Faster-RCNN. This part is trained on 40\% of the images from the CPRI dataset by minimizing binary cross-entropy and smooth $\ell_1$ loss using Adam optimizer with learning rate $10^{-5}$. It is tested on 30\% of CPRI samples for which ground-truths (bounding boxes around ROIs) are generated and not used for training.
	
	\item \textbf{Image Classifier:}
	
	As each image in the datasets used either contains healthy leaf(ves) or one type of disease spot(s), class decision for a whole image is taken based on its detected ROIs. This part is tested on 60\% of the images from CPRI dataset that are not used for training and all of the PV images. 
	 
\end{enumerate}
To evaluate both the ROI detector and image classifier, only predicted ROIs with confidence score more than 80\% are considered.

\subsection{Attention-based Image Classification}

The attention branch network (ABN) \cite{fukui2019attention} consists of three units - feature extractor module, attention branch, and perception branch. The feature extractor contains a stack of convolution layers and extracts a set of feature maps from an input image. We consider ResNet18 as the baseline model for ABN. The attention branch takes these feature maps as input and computes the attention maps using class activation mapping (CAM) \cite{zhou2016learning}  followed by a sigmoid function during training. Attention maps show the location of the highest activated region corresponding to each class. The final probability of each class is computed by the perception branch, which receives the feature map from the feature extractor and the attention map from the attention branch.

The ABN is trained using segmented CPRI images with class labels. Segmented images are generated by training the UNet using manually generated binary masks of a small set of CPRI images. Using the trained UNet, segmentation masks for each of the training and test images from the CPRI dataset are generated, which are further refined using morphological operations. Along with classification scores, ABN also provides the visual cue of the presence of disease spots in images via an attention map, which validates its classification decisions.

\begin{table*}[t]
	\fontsize{9}{11}\selectfont
	\caption{Evaluation metrics of ROI-based image classifier.}
	\label{table:metricsObjDet}
	\centering
	\begin{tabular}{|c|c|c|c||c|c|c|}\hline
		& \multicolumn{3}{|c|}{Overlap Precision (\%)} & \multicolumn{3}{|c|}{Overlap Recall (\%)} \\\hline
		ROI Detector & \multicolumn{3}{|c|} {91.85} &  \multicolumn{3}{|c|} {83.06} \\\hline
		
		& \multicolumn{6}{|c|}{Accuracy (Precision, Recall, F-measure) (\%)} \\\hline
		& \multicolumn{3}{|c|}{Testing} & \multicolumn{3}{|c|}{Cross-testing} \\\hline
		Image Classifier & \multicolumn{3}{|c|}{{96.95} (0.8351, 0.9369, 0.8745)}
		& \multicolumn{3}{|c|}{84.76 (0.7486, 0.8144, 0.7655)} \\\hline
		
		\hline \hline
		
		\multicolumn{7}{|c|}{\textbf{Confusion Matrices}} \\\hline
		\textit{Actual} & \multicolumn{6}{|c|}{\textit{Predicted}} \\\hline	 
		& \multicolumn{3}{|c|}{\textbf{Testing}} & \multicolumn{3}{|c|}{\textbf{Cross-Testing}}  \\\hline
		\multirow{4}{*}{} & Early Blight & Late Blight & Healthy Leaves & Early Blight & Late Blight & Healthy Leaves \\\hline
		Early Blight & \textbf{1197} & 27 & 9 & \textbf{908} & 38 & 54 \\\hline
		Late Blight & 16 & \textbf{753} & 7 & 98 & \textbf{805} & 97 \\\hline
		Healthy Leaves & 3 & 0 & \textbf{20} & 41 & 0 & \textbf{111}\\\hline
		
	\end{tabular}
\end{table*}

\begin{table*}[t]
	\fontsize{9}{11}\selectfont
	\caption{Evaluation metrics of attention-based image classifier.}
	\label{table:metricsAttention}
	\centering
	\begin{tabular}{|c|c|c|c||c|c|c|}\hline
		
		& \multicolumn{6}{|c|}{Accuracy (Precision, Recall, F-measure) (\%)} \\\hline
		& \multicolumn{3}{|c|}{Testing} & \multicolumn{3}{|c|}{Cross-testing} \\\hline
		Image Classifier & \multicolumn{3}{|c|}{94.17 (0.8382, 0.9418, 0.8870)}
		& \multicolumn{3}{|c|}{85.66 (0.7767, 0.8272, 0.8012)} \\\hline
		
		\hline \hline
		
		\multicolumn{7}{|c|}{Confusion Matrices} \\\hline
		Actual & \multicolumn{6}{|c|}{Predicted} \\\hline
		& \multicolumn{3}{|c|}{Testing} & \multicolumn{3}{|c|}{Cross-Testing}  \\\hline
		\multirow{4}{*}{} & Early Blight & Late Blight & Healthy Leaves & Early Blight & Late Blight & Healthy Leaves \\\hline
		Early Blight & 1023 & 22 & 12 & 865 & 3 & 21 \\\hline
		Late Blight &66  & 621 & 2 & 102 & 551 & 92 \\\hline
		Healthy Leaves & 0 & 1 & 22 & 35 & 1 & 102 \\\hline
		
	\end{tabular}
\end{table*}

\begin{figure}[t]
	\centering
	\begin{subfigure}[t]{\dimexpr0.15\textwidth}
		\includegraphics[width=\textwidth]  		{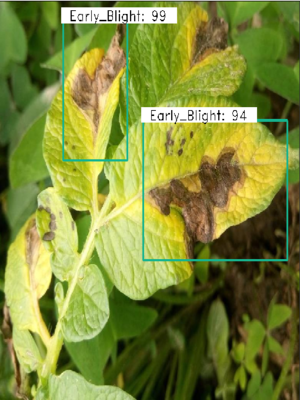}\vspace{0.05cm}
		\includegraphics[width=\textwidth]
{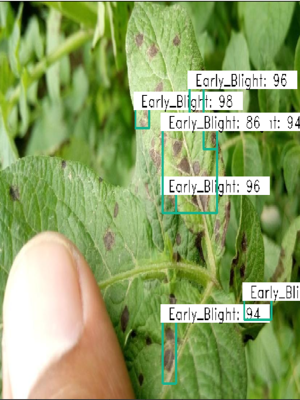}\vspace{0.05cm}
		\includegraphics[width=\textwidth]
		{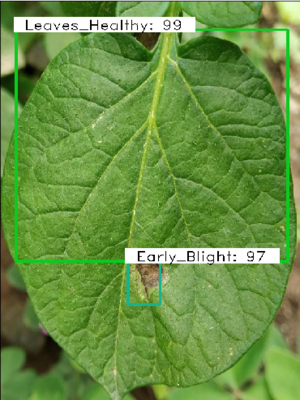}\vspace{0.05cm}
		\caption{EB}
	\end{subfigure}\hspace{0.05cm}
	\begin{subfigure}[t]{0.15\textwidth}
		\includegraphics[width=\textwidth]  
		{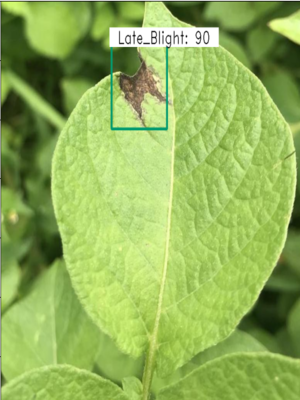}\vspace{0.05cm}
		\includegraphics[width=\textwidth]
		{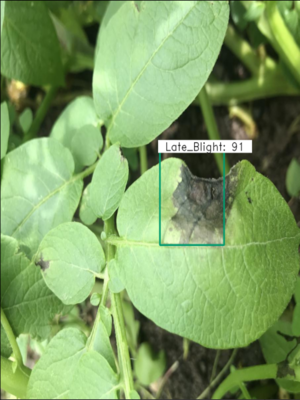}\vspace{0.05cm}
		\includegraphics[width=\textwidth]
		{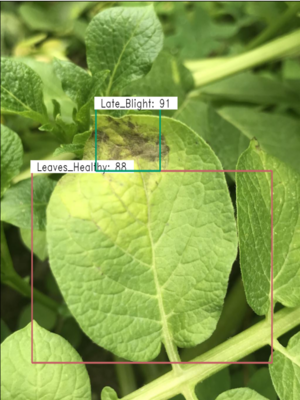}\vspace{0.05cm}
		\caption{LB}
	\end{subfigure}\hspace{0.05cm}
	\begin{subfigure}[t]{0.15\textwidth}
		\includegraphics[width=\textwidth]  
		{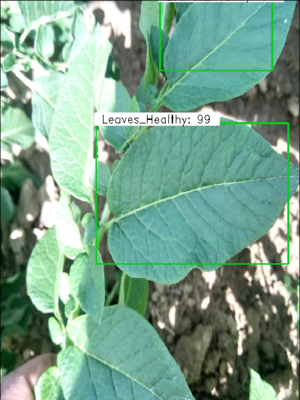}\vspace{0.05cm} 
		\includegraphics[width=\textwidth]  
		{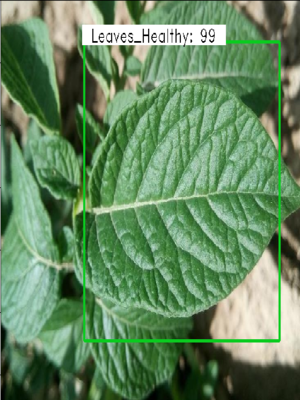}\vspace{0.05cm}
		\includegraphics[width=\textwidth]
		{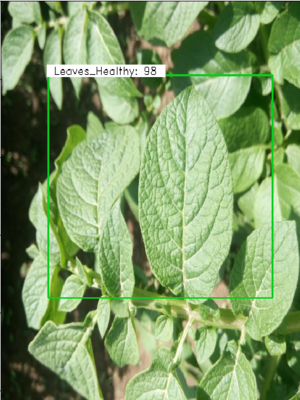}\vspace{0.05cm}	
		\caption{HL}
	\end{subfigure}	
	
	\caption{Sample ROI detection results for original CPRI images. *(\textit{EB: Early Blight; LB: Late Blight; HL: Healthy Leaves}.)}
	\label{fig:objDet_test}
\end{figure}
\begin{figure}[t]
	\centering
	
	\begin{subfigure}[t]{\dimexpr0.17\textwidth}
		\includegraphics[width=\textwidth]  
		{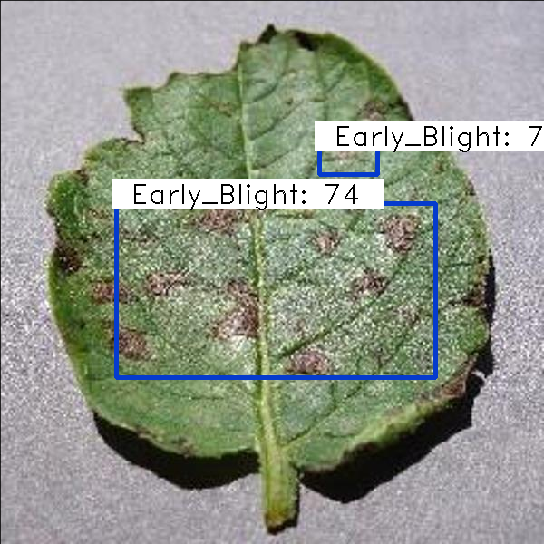}\vspace{0.05cm}
		\includegraphics[width=\textwidth]
		{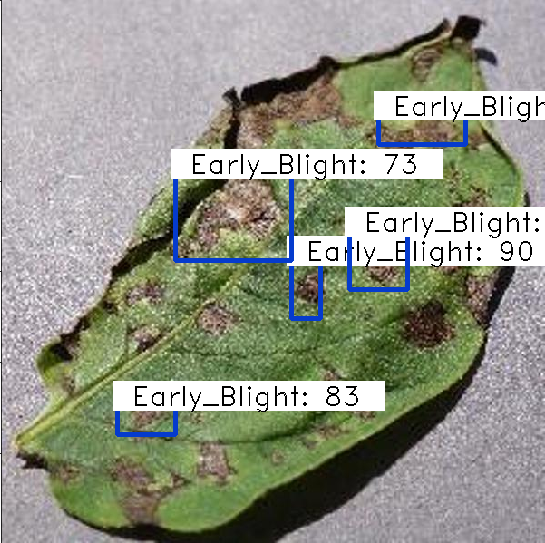}\vspace{0.05cm}
		\includegraphics[width=\textwidth]
		{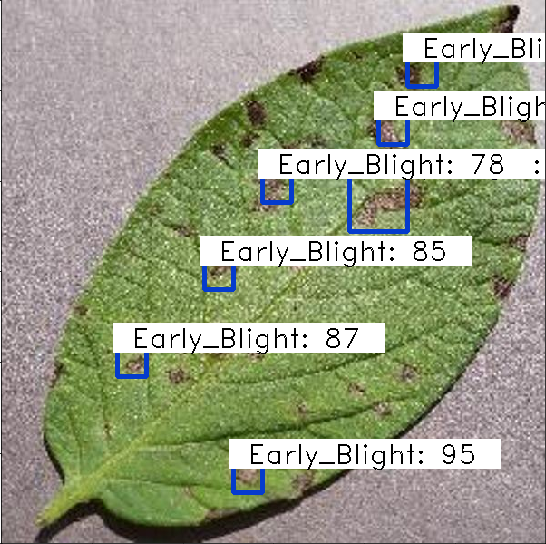}\vspace{0.05cm}
		\caption{EB}
	\end{subfigure}\hspace{0.05cm}
	\begin{subfigure}[t]{0.17\textwidth}
		\includegraphics[width=\textwidth]  
		{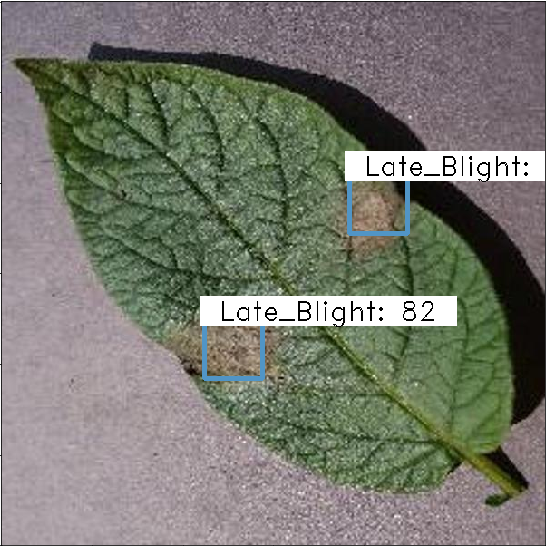}\vspace{0.05cm}
		\includegraphics[width=\textwidth]
		{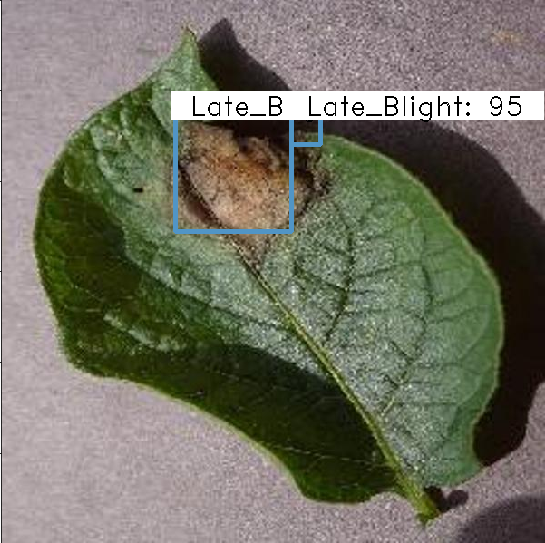}\vspace{0.05cm}
		\includegraphics[width=\textwidth]
		{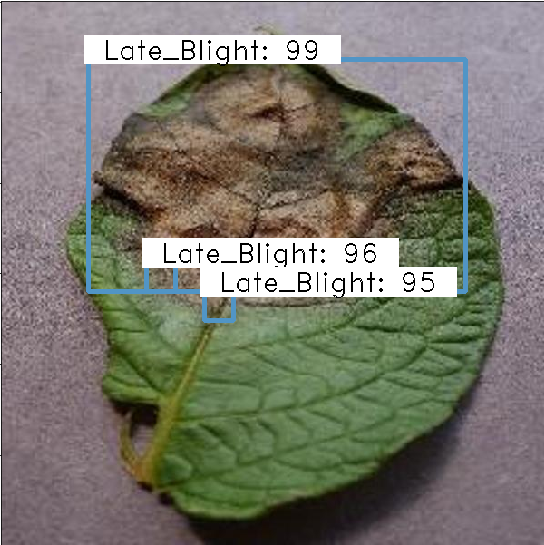}\vspace{0.05cm}
		\caption{LB}
	\end{subfigure}\hspace{0.05cm}
	\begin{subfigure}[t]{0.17\textwidth}
		\includegraphics[width=\textwidth]  
		{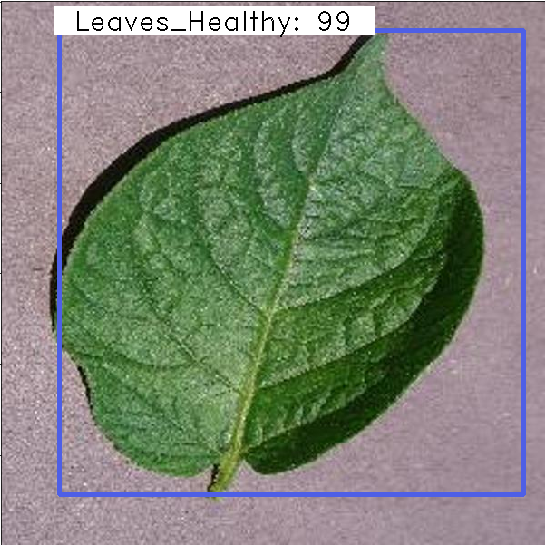}\vspace{0.05cm} 
		\includegraphics[width=\textwidth]  
		{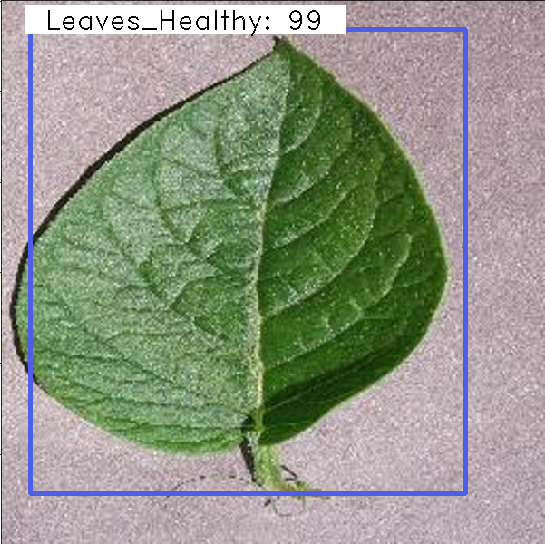}\vspace{0.05cm}
		\includegraphics[width=\textwidth]
		{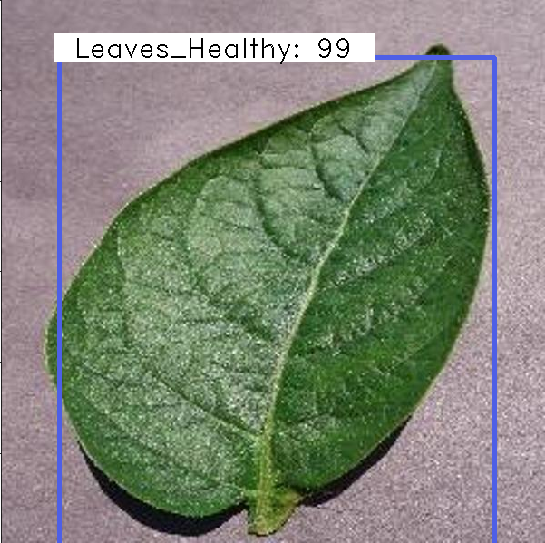}\vspace{0.05cm}	
		\caption{HL}
	\end{subfigure}	
	
	\caption{Sample ROI detection results for PV images. *(\tiny{EB: Early Blight; LB: Late Blight; HL: Healthy Leaves}.)}
	\label{fig:objDet_crossTest}
\end{figure} 
\section{Experimental Results and Analysis} \label{sec:obsAnl}
It can be observed in Figure \ref{fig:ClassSaliency_FTB5} that in saliency maps of any of the datasets, disease spots are not particularly highlighted as the salient regions. Though these spots are quite prominent, the trained models do not concentrate on these for the class decisions. This is proved to be a general trend through the quantitative measures of the correctness of class-specific saliency regions (Table \ref{table:metricsSalReg}). The models, trained on any dataset, exhibit low overlap precision and recall values of salient maps' coverage for the disease classes. As the healthy parts of leaves cover relatively more area in the images, the precision of salient maps for \textit{healthy leaves} class is higher than the disease classes. However, low overlap recall of saliency maps for this class samples in all the datasets indicates that retrieved information is insufficient. Similarly, the foregrounds of disease classes in PV images occupy more area than CPRI images, particularly for original CPRI images, overlapping, leading to the high precision of saliency maps for the PV disease classes. 

In summary, the average overlap precision of the salient maps generated by the models trained on the used datasets is approximately 50\%, which indicates the significant contribution of image background in the learned features. At the same time, less than 30\% overlap recall indicates the insignificance of disease spots in the learned features. Probably, in each of the datasets, a class is attributed to a particular background-foreground combination that exists in the majority of samples from that class. As such combinations are bound to vary with datasets, learned features are not adaptable to the images from similar but different datasets.

These observations indicate that the CNN models trained on the whole leaf images are unable to locate the region of interest (disease spot(s) or healthy leaf) in the images. The trained models fail to segregate the foregrounds of the images from the backgrounds. The following inherent challenges of leaf images account for this phenomenon:  
\begin{inparaenum}
	\item Both the foreground and background in an \textit{in-field} leaf image are of similar appearance - the foreground is a leaf with or without disease spot/s, and the background is other leaves and weeds with exposed ground patches.
	\item Competitive class objects are not quite different in visual appearance from each other.
	\item  Disease spots cover smaller areas of the leaves and are mainly different in colour from the leaves. This colour change seems to be equivalent to the illumination variation in images leading to misinterpretation of image parts by the models \cite{myPaper}.
\end{inparaenum}

Therefore, an effective leaf image-based plant disease classifier must do two tasks simultaneously - differentiate between foreground and background and recognize healthy leaf and two types of disease spots. Hence, the appropriate model for this task must learn to locate the region of interest (ROIs) automatically and must learn features specific to those regions only instead of the whole image for classification. 

The proposed ROI-based and attention-based image classifiers fulfill this need by successfully detecting or highlighting the intended ROIs in an image and thus classifying the whole image correctly (Table \ref{table:metricsObjDet} and Table \ref{table:metricsAttention}). The ROI-based classifier detected ROIs in images with 91.85\% precision and its testing accuracy is 96.93\%. These high metrics prove that the proposed classifier learned the true features to differentiate between background and foreground and between the class-specific ROIs. The equivalent \textit{cross-testing} accuracy (84.76\%) proves that the classifier is independent of the environment of captured images and is adaptable. 

The ROI-based classifier also detected most of the desired ROIs as the recall of ground-truth overlap is 83.06\%. Less than 10\% of the missed detections are misclassified ROIs. The missed detections are either very small disease spots or are over or underestimation of ground-truth boxes due to the arbitrary shapes of disease spots. Figure \ref{fig:objDet_test} and Figure \ref{fig:objDet_crossTest} show some of the results of the ROI detector. 

\begin{figure}[t]
	\centering
 \begin{subfloat}[Early Blight]
		{\includegraphics[width=2cm]  
			{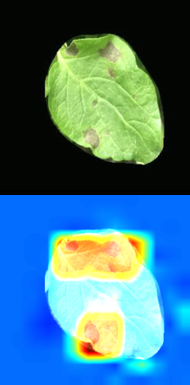}
			\includegraphics[width=2cm]
			{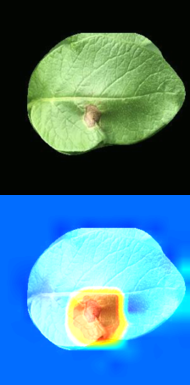}
			\includegraphics[width=2cm]
			{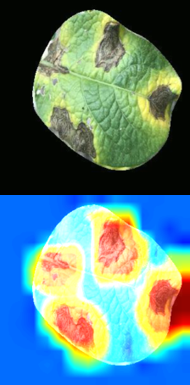}}
	\end{subfloat}\vspace{0.2cm}
	\begin{subfloat}[Late Blight]
		{\includegraphics[width=2cm]  
			{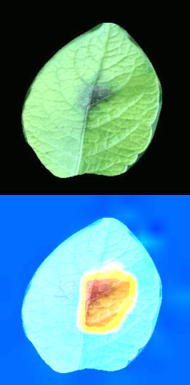}
			\includegraphics[width=2cm]
			{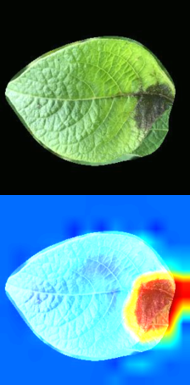}
			\includegraphics[width=2cm]
			{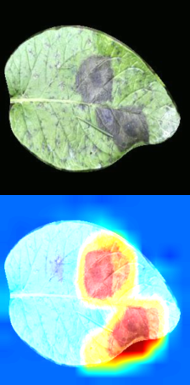}}
	\end{subfloat}\vspace{0.2cm}
	\begin{subfloat}[Healthy Leaves]
		{\includegraphics[width=2cm]  
			{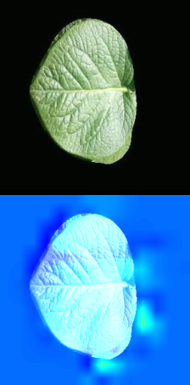}
			\includegraphics[width=2cm]
			{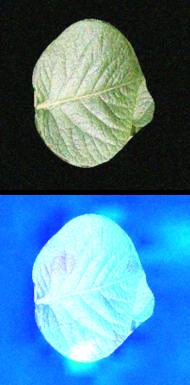}
			\includegraphics[width=2cm]
			{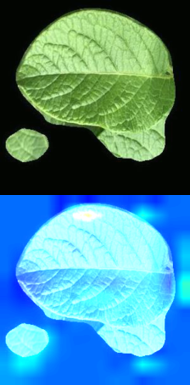}}
	\end{subfloat}
	
	\caption{Visualisation of region of interest in CPRI images through ABN network.}
	\label{fig:abn_vis_CPRI}
\end{figure}  

\begin{figure}[t]
	\centering
	
	\begin{subfloat}[Early Blight]
		{\includegraphics[width=2cm]  
			{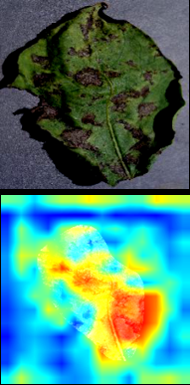}
			\includegraphics[width=2cm]
			{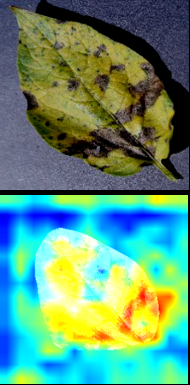}
			\includegraphics[width=2cm]
			{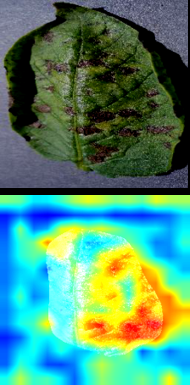}}
	\end{subfloat}\vspace{0.2cm}
	\begin{subfloat}[Late Blight]
		{\includegraphics[width=2cm]  
			{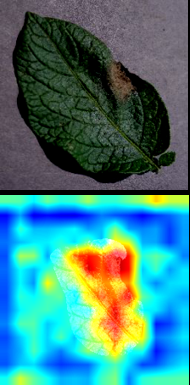}
			\includegraphics[width=2cm]
			{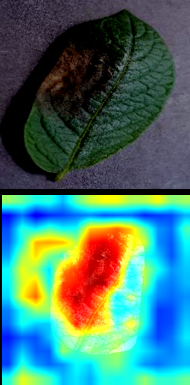}
			\includegraphics[width=2cm]
			{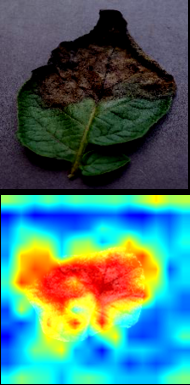}}
	\end{subfloat}\vspace{0.2cm}
	\begin{subfloat}[Healthy Leaves]
		{\includegraphics[width=2cm]  
			{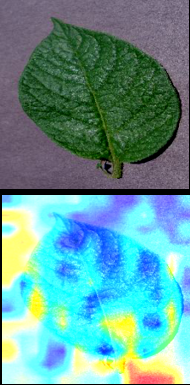}
			\includegraphics[width=2cm]
			{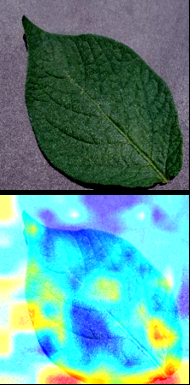}
			\includegraphics[width=2cm]
			{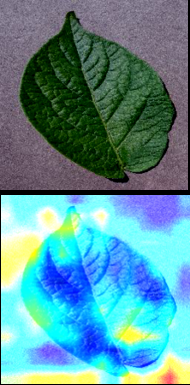}}
	\end{subfloat}
	
	\caption{Visualisation of the region of interest in PV images through ABN network.}
	\label{fig:abn_vis_PV}
\end{figure}  
ABN also considers the disease regions for classification decisions for images of both CPRI and PV datasets, as seen in Figure \ref{fig:abn_vis_CPRI} and Figure \ref{fig:abn_vis_PV}. It is observed that for disease class images, the attention map highlights the disease areas. In contrast, there is no attention region in the images of \textit{healthy leaves} class, as these images do not contain any disease spot. For the ability to detect attention regions, ABN classified CPRI images with 94.95\% accuracy. Its 85.66\% cross-testing accuracy also proves that ABN learned features from diseased areas and healthy leaf.

\section{Conclusion} \label{sec:conc}

Our experimental observations show that for plant leaf disease detection, the conventional CNNs are unable to detect the region of interest in the images and learn discriminative features from the overall image content of their respective training datasets. Hence, these features are not precisely related to the disease spot(s) and healthy leaf and are not applicable to other similar potato leaf image datasets. 

On the other hand, the high classification accuracies of the proposed classifiers prove that ``region-specific feature learning" suppressed the backgrounds and the visual similarities between the objects of different classes. The equivalent \textit{testing and cross-testing} results prove that the region and attention-based classifiers are independent of the environment of captured images and are adaptable to different but similar datasets. 

However, the effectiveness of such systems can be ensured by practicing some precautionary measures while capturing the images, which cannot be addressed completely by algorithms - \begin{inparaenum}[1)]
	\item regions of interest, particularly the disease spots, must be focused, 
	\item regions of interest must be evenly illuminated,
	\item training data must represent the possible intra-class variations, highlighting the inter-class variations. 
\end{inparaenum}  
\section*{Acknowledgments}
This research work is a part of the project \textit{FarmerZone}, sponsored by the Department of Biotechnology, Govt. of India. We thank Central Potato Research Institute (CPRI), Shimla, India, for providing the dataset from the field.

\printbibliography

\end{document}